\def\year{2018}\relax
\documentclass[letterpaper]{article} 
\usepackage{aaai18}  
\usepackage{times}  
\usepackage{helvet}  
\usepackage{courier}  
\usepackage{url}  
\usepackage{graphicx}  
\usepackage{subcaption}
\usepackage{courier}
\usepackage{multirow}
\usepackage{color}
\usepackage{todonotes}
\usepackage{amssymb}
\usepackage{multirow}
\usepackage[linesnumbered, ruled]{algorithm2e}
\usepackage{amsmath}

\begin{document}
%
\title{Deep Multi-View Spatial-Temporal Network for Taxi Demand Prediction}
\author{Huaxiu Yao\thanks{The paper was done when these authors were interns in Didi Chuxing.}, Fei Wu\\Pennsylvania State University\\\{huaxiuyao, fxw133\}@ist.psu.edu \And Jintao Ke\footnotemark[1]\\Hong Kong University of Science \\and Technology\\jke@connect.ust.hk \And Xianfeng Tang\\Pennsylvania State University\\xianfeng@ist.psu.edu \AND Yitian Jia, Siyu Lu, Pinghua Gong, Jieping Ye\\Didi Chuxing\\\{jiayitian, lusiyu, gongpinghua, yejieping\}@didichuxing.com \And Zhenhui Li\\Pennsylvania State University\\jessieli@ist.psu.edu}

\maketitle
\begin{abstract}
	Taxi demand prediction is an important building block to enabling intelligent transportation systems in a smart city. An accurate prediction model can help the city pre-allocate resources to meet travel demand and to reduce empty taxis on streets which waste energy and worsen the traffic congestion.  With the increasing popularity of taxi requesting services such as Uber and Didi Chuxing (in China), we are able to collect large-scale taxi demand data continuously. How to utilize such big data to improve the demand prediction is an interesting and critical real-world problem. Traditional demand prediction methods mostly rely on time series forecasting techniques, which fail to model the complex non-linear spatial and temporal relations. Recent advances in deep learning have shown superior performance on traditionally challenging tasks such as image classification by learning the complex features and correlations from large-scale data. This breakthrough has inspired researchers to explore deep learning techniques on traffic prediction problems. However, existing methods on traffic prediction have only considered spatial relation (e.g., using CNN) or temporal relation (e.g., using LSTM) independently. We propose a Deep Multi-View Spatial-Temporal Network (DMVST-Net) framework to model both spatial and temporal relations. Specifically, our proposed model consists of three views: temporal view (modeling correlations between future demand values with near time points via LSTM), spatial view (modeling local spatial correlation via local CNN), and semantic view (modeling correlations among regions sharing similar temporal patterns). Experiments on large-scale real taxi demand data demonstrate effectiveness of our approach over state-of-the-art methods.
\end{abstract}

\section{Introduction}
Traffic is the pulse of a city that impacts the daily life of millions of people. One of the most fundamental questions for future smart cities is how to build an efficient transportation system. To address this question, a critical component is an accurate demand prediction model. The better we can predict demand on travel, the better we can pre-allocate resources to meet the demand and avoid unnecessary energy consumption. Currently, with the increasing popularity of taxi requesting services such as Uber and Didi Chuxing, we are able to collect massive demand data at an unprecedented scale. The question of how to utilize big data to better predict traffic demand has drawn increasing attention in AI research communities. 

In this paper, we study the taxi demand prediction problem; that problem being how to predict the number of taxi requests for a region in a future timestamp by using historical taxi requesting data. In literature, there has been a long line of studies in traffic data prediction, including traffic volume, taxi pick-ups, and traffic in/out flow volume. To predict traffic, time series prediction methods have frequently been used. Representatively, autoregressive integrated moving average (ARIMA) and its variants have been widely applied for traffic prediction~\cite{li2012prediction,moreira2013predicting,shekhar2008adaptive}. Based on the time series prediction method, recent studies further consider spatial relations~\cite{deng2016latent,tong2017sim} and external context data (e.g., venue, weather, and events)~\cite{pan2012utilizing,wu2016interpreting}. While these studies show that prediction can be improved by considering various additional factors, they still fail to capture the complex nonlinear spatial-temporal correlations.

Recent advances in deep learning have enabled researchers to model the complex nonlinear relationships and have shown promising results in computer vision and natural language processing fields~\cite{lecun2015deep}. This success has inspired several attempts to use deep learning techniques on traffic prediction problems. Recent studies \cite{zhang2016deep,zhang2016dnn} propose to treat the traffic in a city as an image and the traffic volume for a time period as pixel values. Given a set of historical traffic images, the model predicts the traffic image for the next timestamp. Convolutional neural network (CNN) is applied to model the complex spatial correlation. \citeauthor{yu2017deep}~\shortcite{yu2017deep} proposes to use Long Short Term Memory networks (LSTM) to predict loop sensor readings.
They show the proposed LSTM model is capable of modeling complex sequential interactions.
These pioneering attempts show superior performance compared with previous methods based on traditional time series prediction methods. However, none of them consider spatial relation and temporal sequential relation simultaneously.

In this paper, we harness the power of CNN and LSTM in a joint model that captures the complex nonlinear relations of both space and time. However, we cannot simply apply CNN and LSTM on demand prediction problem. If treating the demand over an entire city as an image and applying CNN on this image, we fail to achieve the best result. We realize including regions with weak correlations to predict a target region actually hurts the performance. To address this issue, we propose a novel local CNN method which only considers spatially nearby regions. This local CNN method is motivated by the First Law of Geography: ``near things are more related than distant things,''~\cite{tobler1970computer} and it is also supported by observations from real data that demand patterns are more correlated for spatially close regions.

While local CNN method filters weakly correlated remote regions, this fails to consider the case that two locations could be spatially distant but are similar in their demand patterns (i.e., on the semantic space). For example, residential areas may have high demands in the morning when people transit to work, and commercial areas may be have high demands on weekends. We propose to use a graph of regions to capture this latent semantic, where the edge represents similarity of demand patterns for a pair of regions. Later, regions are encoded into vectors via a graph embedding method and such vectors are used as context features in the model. In the end, a fully connected neural network component is used for prediction.

Our method is validated via large-scale real-world taxi demand data from Didi Chuxing. The dataset contains taxi demand requests through Didi service in the city of Guangzhou in China over a two-month span, with about 300,000 requests per day on average. We conducted extensive experiments to compare with state-of-the-art methods and have demonstrated the superior performance of our proposed method.

In summary, our contributions are summarized as follow:
\begin{itemize}
	\item We proposed a unified multi-view model that jointly considers the spatial, temporal, and semantic relations.
	\item We proposed a local CNN model that captures local characteristics of regions in relation to their neighbors. 
	\item We constructed a region graph based on the similarity of demand patterns in order to model the correlated but spatially distant regions. The latent semantics of regions are learnt through graph embedding.
	\item  We conducted extensive experiments on a large-scale taxi request dataset from Didi Chuxing. The results show that our method consistently outperforms the competing baselines. 
	
\end{itemize}

\section{Related Work}
Problems of traffic prediction could include predicting any traffic related data, such as traffic volume (collected from GPS or loop sensors), taxi pick-ups or drop-offs,  traffic flow, and taxi demand (our problem). The problem formulation process for these different types of traffic data is the same. Essentially, the aim is to predict a traffic-related value for a location at a timestamp. In this section, we will discuss the related work on traffic prediction problems.

The traditional approach is to use time series prediction method. Representatively, autoregressive integrated moving average (ARIMA) and its variants have been widely used in traffic prediction problem~\cite{shekhar2008adaptive,li2012prediction,moreira2013predicting}. 

Recent studies further explore the utilities of external context data, such as venue types, weather conditions, and event information~\cite{pan2012utilizing,wu2016interpreting,wang2017deepsd,tong2017sim}. In addition, various techniques have also been introduced to model spatial interactions. For example,~\citeauthor{deng2016latent}~\shortcite{deng2016latent} used matrix factorization on road networks to capture a correlation among road connected regions for predicting traffic volume. Several studies~\cite{tong2017sim,ide2011trajectory,zheng2013time} also propose to smooth the prediction differences for nearby locations and time points via regularization for close space and time dependency. These studies assume traffic in nearby locations should be similar. However, all of these methods are based on the time series prediction methods and fail to model the complex nonlinear relations of the space and time.

Recently, the success of deep learning in the fields of computer vision and natural language processing~\cite{lecun2015deep,krizhevsky2012imagenet} motivates researchers to apply deep learning techniques on traffic prediction problems. For instance, \citeauthor{wang2017deepsd}~\shortcite{wang2017deepsd} designed a neural network framework using context data from multiple sources and predict the gap between taxi supply and demand.
The method uses extensive features, but does not model the spatial and temporal interactions.

A line of studies applied CNN to capture spatial correlation by treating the entire city's traffic as images. For example, \citeauthor{ma2017learning}~\shortcite{ma2017learning} utilized CNN on images of traffic speed for the speed prediction problem. \citeauthor{zhang2016dnn}~\shortcite{zhang2016dnn} and \citeauthor{zhang2016deep}~\shortcite{zhang2016deep} proposed to use residual CNN on the images of traffic flow. These methods simply use CNN on the whole city and will use all the regions for prediction. We observe that utilizing irrelevant regions (e.g., remote regions) for prediction of the target region might actually hurts the performance. In addition, while these methods do use traffic images of historical timestamps for prediction, but they do not explicitly model the temporal sequential dependency. 

Another line of studies uses LSTM for modeling sequential dependency. 
\citeauthor{yu2017deep}~\shortcite{yu2017deep} proposed to apply Long-short-term memory (LSTM) network and autoencoder to capture the sequential dependency for predicting the traffic under extreme conditions, particularly for peak-hour and post-accident scenarios. However, they do not consider the spatial relation. 

In summary, the biggest difference of our proposed method compared with literature is that we consider \emph{both} spatial relation and temporal sequential relation in a joint deep learning model.
\section{Preliminaries}
In this section, we first fix some notations and define the taxi demand problem. We follow previous studies~\cite{zhang2016deep,wang2017deepsd} and define the set of non-overlapping locations $L = \{l_1,l_2,..., l_i, ... ,l_N\}$ as rectangle partitions of a city,
and the set of time intervals as $\mathcal{I} = \{I_0, I_1, ..., I_t, ..., I_T\}$. 30 minutes is set as the length of the time interval. Alternatively, more sophisticated ways of partitioning can also be used, such as partition space by road network~\cite{deng2016latent} or hexagonal partitioning. However, this is not the focus of this paper, and our methodology can still be applied. Given the set of locations $L$ and time intervals $T$, we further define the following.

\textbf{Taxi request}: 
A taxi request $o$ is defined as a tuple $(o.t, o.l, o.u)$, where $o.t$ is the timestamp, $o.l$ represents the location, and $o.u$ is user identification number. The requester identifications are used for filtering duplicated and spammer requests.

\textbf{Demand}:
The demand is defined as the number of taxi requests at one location per time point, i.e., 
$y_t^i = |\{o : o.t\in I_t \wedge o.l\in l_i \}| $, where $|\cdot|$ denotes the cardinality of the set. For simplicity, we use the index of time intervals $t$ representing $I_t$, and the index of locations $i$ representing $l_i$ for rest of the paper.

\textbf{Demand prediction problem}:
The demand prediction problem aims to predict the demand at time interval $t+1$, given the data until time interval $t$. In addition to historical demand data, we can also incorporate context features such as temporal features, spatial features, meteorological features (refer to Data Description section for more details). We denote those context features for a location $i$ and a time point $t$ as a vector $ \mathbf{e}^i_t \in \mathbb{R}^r$, where $r$ is the number of features. Therefore, our final goal is to predict 
\begin{equation*}
y_{t+1}^i = \mathcal{F}(\mathcal{Y}^L_{t-h,...,t}, \mathcal{E}^L_{t-h,...,t})
\end{equation*} 
for $i \in L$, where $\mathcal{Y}^L_{t-h,...,t}$ are historical demands and $\mathcal{E}^L_{t-h,...,t}$ are context features for all locations $L$ for time intervals from $t-h$ to $t$, where $t-h$ denotes the starting time interval. We define our prediction function $\mathcal{F}(\cdot)$ on all regions and previous time intervals up to $t-h$ to capture the complex spatial and temporal interaction among them. 

\section{Proposed DMVST-Net Framework}

\begin{figure*}[t]
	\centering
	\includegraphics[width=0.82\textwidth]{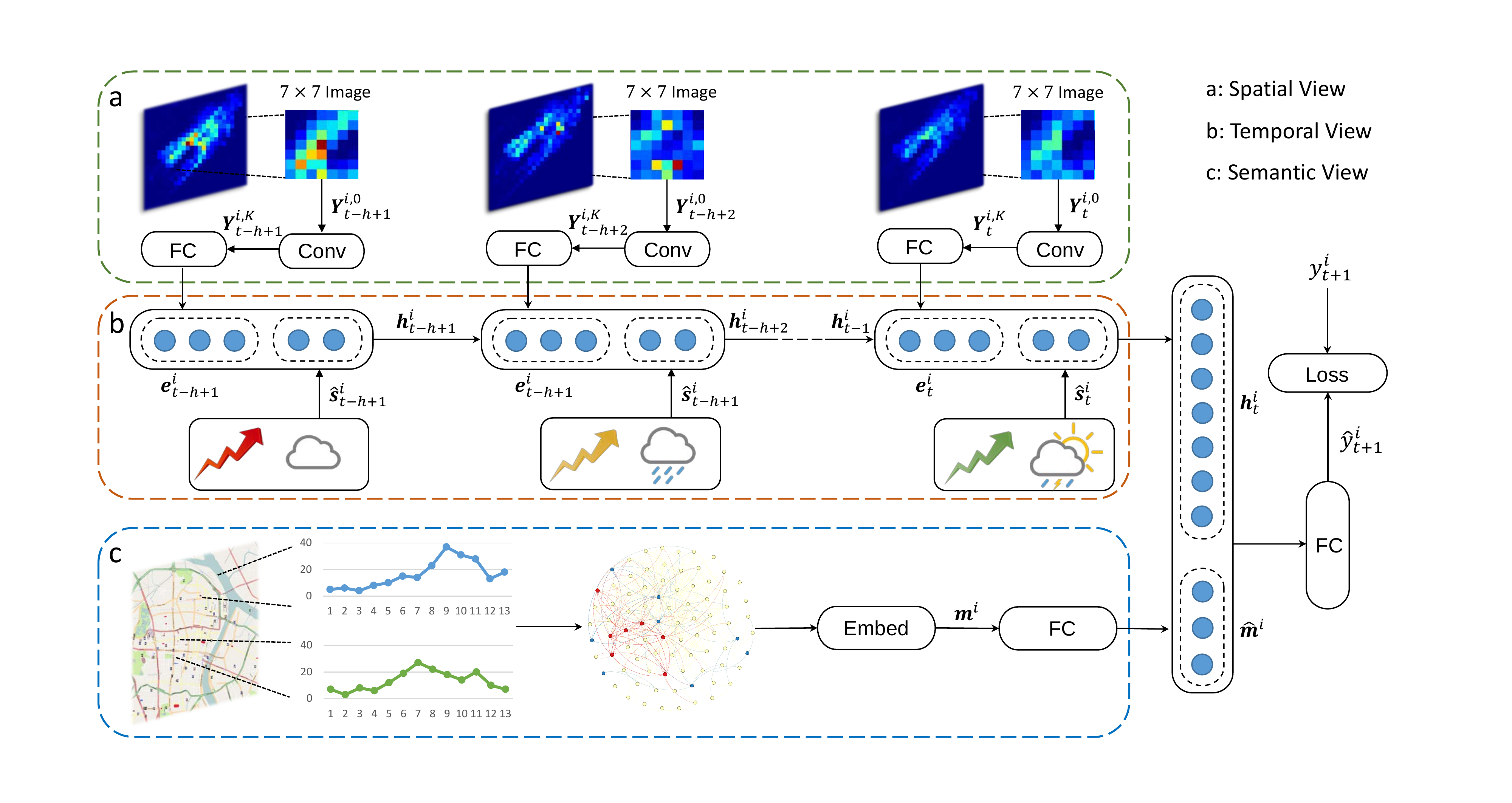}
	\caption{The Architecture of DMVST-Net. 
		(a). The spatial component uses a local CNN to capture spatial dependency among nearby regions. The local CNN includes several convolutional layers. A fully connected layer is used at the end to get a low dimensional representation. (b). The temporal view employs a LSTM model, which takes the representations from the spatial view and concatenates them with context features at corresponding times. 
		(c). The semantic view first constructs a weighted graph of regions (with weights representing functional similarity). Nodes are encoded into vectors. A fully connected layer is used at the end for jointly training. Finally, a fully connected neural network is used for prediction.}
	\label{fig:framework}
\end{figure*}

In this section, we provide details for our proposed Deep Multi-View Spatial-Temporal Network (DMVST-Net) framework, i.e., our prediction function $\mathcal{F}$. Figure~\ref{fig:framework} shows the architecture of our proposed method. Our proposed model has three views: spatial, temporal, and semantic view.

\subsection{Spatial View: Local CNN}
As we mentioned earlier, including regions with weak correlations to predict a target region actually hurts the performance. To address this issue, we propose a local CNN method which only considers spatially nearby regions. Our intuition is motivated by the First Law of Geography~\cite{tobler1970computer} - ``near things are more related than distant things''. 

As shown in Figure~\ref{fig:framework}(a), at each time interval $t$, we treat one location $i$ with its surrounding neighborhood as one $S \times S$ image (e.g., $7\times 7$ image in Figure~\ref{fig:framework}(a)) having one channel of demand values (with $i$ being at the center of the image), where the size $S$ controls the spatial granularity. We use zero padding for location at boundaries of the city. As a result, we have an image as a tensor (having one channel) $\mathbf{Y}_t^{i} \in \mathbb{R}^{S\times S\times 1}$, for each location $i$ and time interval $t$. The local CNN takes $\mathbf{Y}_t^{i}$ as input $\mathbf{Y}_t^{i,0}$ and feeds it into $K$ convolutional layers. The transformation at each layer $k$ is defined as follows:
\begin{equation}
\mathbf{Y}_t^{i,k}=f(\mathbf{Y}_t^{i,k-1}\ast \mathbf{W}_t^{k} +\mathbf{b}_t^{k}),
\end{equation}
where $\ast$ denotes the convolutional operation and $f(\cdot)$ is an activation function. In this paper, we use the rectifier function as the activation, i.e., $f(z)=max(0, z)$. $\mathbf{W}_t^{k}$ and $\mathbf{b}_t^{k}$ are two sets of parameters in the $k^{th}$ convolution layer. Note that the parameters $\mathbf{W}_t^{1,...,K}$ and $\mathbf{b}_t^{1,...,K}$ are shared across all regions $i \in L$ to make the computation tractable.

After $K$ convolution layers, we use a flatten layer to transform the output $\mathbf{Y}_t^{i,K}\in \mathbb{R}^{ S \times S \times \lambda}$ to a feature vector $\mathbf{s}^i_t \in \mathbb{R}^{ S^2 \lambda}$ for region $i$ and time interval $t$. At last, we use a fully connected layer to reduce the dimension of spatial representations  $\mathbf{s}^i_t$, which is defined as: 
\begin{equation}
\hat{\mathbf{s}}^i_t=f(W^{fc}_t\mathbf{s}^i_t+b^{fc}_t),
\end{equation}
where $W^{fc}_t$ and $b^{fc}_t$ are two learnable parameter sets at time interval $t$. Finally, for each time interval $t$, we get the $\hat{\mathbf{s}}_t^i \in \mathbb{R}^d$ as the representation for region $i$.

\subsection{Temporal View: LSTM}
The temporal view models sequential relations in the demand time series. We propose to use Long Short-Term Memory (LSTM) network as our temporal view component. LSTM~\cite{hochreiter1997long} is a type of neural network structure, which provides a good way to model sequential dependencies by recursively applying a transition function to the hidden state vector of the input. It is proposed to address the problem of classic Recurrent Neural Network~(RNN) for its exploding or vanishing of gradient in the long sequence training~\cite{hochreiter2001gradient}.

LSTM learns sequential correlations stably by maintaining a \emph{memory cell} $\mathbf{c}_t$ in time interval $t$, which can be regarded as an accumulation of previous sequential information. In each time interval, LSTM takes an input $\mathbf{g}_t^i$, $\mathbf{h}_{t-1}$ and $\mathbf{c}_{t-1}$ in this work, and then all information is accumulated to the memory cell when the \emph{input gate} $\mathbf{i}^i_t$ is activated. In addition, LSTM has a \emph{forget gate} $\mathbf{f}^i_t$. If the forget gate is activated, the network can forget the previous memory cell $\mathbf{c}^i_{t-1}$. Also, the \emph{output gate} $\mathbf{o}^i_t$ controls the output of the memory cell. In this study, the architecture of LSTM is formulated as follows:
\begin{equation}
\begin{aligned}
&\mathbf{i}^i_t=\sigma(\mathbf{W}_i\mathbf{g}_t^i+\mathbf{U}_i\mathbf{h}^i_{t-1}+\mathbf{b}_i),\\
&\mathbf{f}^i_t=\sigma(\mathbf{W}_f\mathbf{g}_t^i+\mathbf{U}_f\mathbf{h}^i_{t-1}+\mathbf{b}_f),\\
&\mathbf{o}^i_t=\sigma(\mathbf{W}_o\mathbf{g}_t^i+\mathbf{U}_o\mathbf{h}^i_{t-1}+\mathbf{b}_o),\\
&\mathbf{\theta}^i_t=\tanh(\mathbf{W}_g\mathbf{g}_t^i+\mathbf{U}_g\mathbf{h}^i_{t-1}+\mathbf{b}_g),\\
&\mathbf{c}^i_t=\mathbf{f}^i_t\circ \mathbf{c}^i_{t-1}+\mathbf{i}^i_t\circ \mathbf{\theta}^i_t,\\
&\mathbf{h}^i_t=\mathbf{o}^i_t\circ \tanh(\mathbf{c}^i_t).\\
\end{aligned}
\end{equation}
where $\circ$ denotes Hadamard product and $\tanh$ is hyperbolic tangent function. Both functions are element-wise. $\mathbf{W}_a, \mathbf{U}_a, \mathbf{b}_a$ ($a\in\{i,f,o,g\}$) are all learnable parameters. The number of time intervals in LSTM is $h$ and the output of region $i$ of LSTM after $h$ time intervals is $\mathbf{h}^i_t$.

As Figure~\ref{fig:framework}(b) shows, the temporal component takes representations from the spatial view and concatenates them with context features. More specifically, we define:
\begin{equation}
\mathbf{g}_t^i=\hat{\mathbf{s}}^i_t \oplus \mathbf{e}^i_t,
\end{equation}
where $\oplus$ denotes the concatenation operator, therefore, $\mathbf{g}_t^i \in \mathbb{R}^{r+d}$.

\subsection{Semantic View: Structural Embedding}
Intuitively, locations sharing similar functionality may have similar demand patterns, e.g., residential areas may have a high number of demands in the morning when people transit to work, and commercial areas may expect to have high demands on weekends. Similar regions may not necessarily be close in space. Therefore, we construct a graph of locations representing functional (semantic) similarity among regions.

We define the semantic graph of location as $G=(V,E,D)$, where the set of locations $L$ are nodes $V=L$, $E \in V\times V$ is the edge set, and $D$ is a set of similarity on all the edges. We use Dynamic Time Warping (DTW) to measure the similarity $\omega_{ij}$ between node (location) $i$ and node (location) $j$. 
\begin{equation}
\omega_{ij} = \exp(-\alpha {\rm DTW}(i, j)),
\end{equation}
where $\alpha$ is the parameter that controls the decay rate of the distance (in this paper, $\alpha=1$), and ${\rm DTW}(i, j)$ is the dynamic time warping distance between the demand patterns of two locations. We use the average weekly demand time series as the demand patterns. The average is computed on the training data in the experiment. The graph is fully connected because every two regions can be reached. 

In order to encode each node into a low dimensional vector and maintain the structural information, we apply a graph embedding method on the graph. For each node $i$ (location), the embedding method outputs the embedded feature vector $\mathbf{m}^i$. In addition, in order to co-train the embedded $\mathbf{m}^i$ with our whole network architecture, we feed the feature vector $\mathbf{m}^i$ to a fully connected layer, which is defined as:
\begin{equation}
\hat{\mathbf{m}}^i=f(W_{fe}\mathbf{m}^i+b_{fe}),
\end{equation}
$W_{fe}$ and $b_{fe}$ are both learnable parameters. In this paper, we use LINE for generating embeddings~\cite{tang2015line}.

\subsection{Prediction Component}
Recall that our goal is to predict the demand at $t+1$ given the data till $t$. We join three views together by concatenating $\hat{\mathbf{m}}^i$ with the output $\mathbf{h}_t^i$ of LSTM:
\begin{equation}
\mathbf{q}_t^i=\mathbf{h}_t^i \oplus \hat{\mathbf{m}}^i.
\end{equation}
Note that the output of LSTM $\mathbf{h}_t^i$ contains both effects of temporal and spatial view. Then we feed $\mathbf{q}_t^i$ to the fully connected network to get the final prediction value $\hat{y}_{t+1}^i$ for each region. We define our final prediction function as:
\begin{equation}
\hat{y}_{t+1}^i=\sigma(W_{ff}\mathbf{q}^i_t+b_{ff}),
\end{equation}
where $W_{ff}$ and $b_{ff}$ are learnable parameters. $\sigma(x)$ is a Sigmoid function defined as $\sigma(x)=1/(1+e^{-x})$. The output of our model is in $[0,1]$, as the demand values are normalized. We later denormalize the prediction to get the actual demand values.

\subsection{Loss function}
In this section, we provide details about the loss function used for jointly training our proposed model. The loss function we used is defined as:
\begin{equation}
\label{eq:loss}
\mathcal{L}(\theta) = \sum_{i=1}^N((y^i_{t+1} - \hat{y}^i_{t+1})^2 + \gamma (\frac{y^i_{t+1} -\hat{y}^i_{t+1}}{y^i_{t+1}})^2),
\end{equation}
where $\theta$ are all learnable parameters in the DMVST-Net and $\gamma$ is a hyper parameter. The loss function consists of two parts: mean square loss and square of mean absolute percentage loss. In practice, mean square error is more relevant to predictions of large values. To avoid the training being dominated by large value samples, we in addition minimize the mean absolute percentage loss. Note that, in the experiment, all compared regression methods use the same loss function as defined in Eq.~\eqref{eq:loss} for fair comparison. The training pipeline is outlined in Algorithm~\ref{alg:outline}. We use Adam~\cite{kingma2014adam} for optimization. We use Tensorflow and Keras~\cite{chollet2015keras} to implement our proposed model.

\begin{algorithm}[t]
	\caption{Training Pipeline of DMVST-Net}
	\label{alg:outline}
	\KwIn{Historical observations: $\mathcal{Y}^L_{1,...,t}$;
		Context features:  $\mathcal{E}^L_{t-h,...,t}$;
		Region structure graph $G=(V,E,D)$;
		Length of the time period $h$;}
	\KwOut{Learned DMVST-Net model}
	
	Initialization\;
	\For{$\forall i \in L$}{
		Use LINE on $G$ and get the embedding result $\mathbf{m}^i$\;
		\For{ $\forall t \in [h, T]$}{
			$\mathcal{S}_{spa}=[\mathbf{Y}_{t-h+1}^i, \mathbf{Y}_{t-h+2}^i,..., \mathbf{Y}_t^i]$\;
			$\mathcal{S}_{cox}=[\mathbf{e}^i_{t-h+1}, \mathbf{e}^i_{t-h+2},..., \mathbf{e}^i_t]$\;
			Append $<\{\mathcal{S}_{spa}$, $\mathcal{S}_{cox}$, $\mathbf{m}^i\},y^i_{t+1}>$ to $\Omega_{bt}$ \;}
	}
	Initialize all learnable parameters $\theta$ in DMVST-Net\;
	\Repeat{stopping criteria is met}{
		Randomly select a batch of instance $\Omega_{bt}$ from $\Omega$\;
		Optimize $\theta$ by minimizing the loss function Eq.~\eqref{eq:loss} with $\Omega_{bt}$
	}
	
\end{algorithm}

\section{Experiment}
\subsection{Dataset Description}
In this paper, we use a large-scale online taxi request dataset collected from Didi Chuxing, which is one of the largest online car-hailing companies in China. The dataset contains taxi requests from $02/01/2017$ to $03/26/2017$ for the city of Guangzhou. 
There are $20\times 20$ regions in our data. The size of each region is $0.7km\times 0.7km$. 
There are about $300,000$ requests each day on average. The context features used in our experiment are the similar types of features used in~\cite{tong2017sim}. These features include temporal features (e.g., the average demand value in the last four time intervals), spatial features (e.g., longitude and latitude of the region center), meteorological features (e.g., weather condition), event features (e.g., holiday).

In the experiment, the data from $02/01/2017$ to $03/19/2017$ is used for training ($47$ days), and the data from $03/20/2017$ to $03/26/2017$ ($7$ days) is used for testing. We use half an hour as the length of the time interval. When testing the prediction result, we use the previous 8 time intervals (i.e., 4 hours) to predict the taxi demand in the next time interval. In our experiment, we filter the samples with demand values less than 10. This is a common practice used in industry. Because in the real-world applications, people do not care about such low-demand scenarios.

\subsection{Evaluation Metric}
We use Mean Average Percentage Error (MAPE) and Rooted Mean Square Error (RMSE) to evaluate our algorithm, which are defined as follows:
\begin{equation}
MAPE=\frac{1}{\xi}\sum_{i=1}^{\xi}\frac{|\hat{y}_{t+1}^i-y_{t+1}^i|}{y_{t+1}^i},
\end{equation}
\begin{equation}
RMSE=\sqrt{\frac{1}{\xi} \sum_{i=1}^\xi(\hat{y}_{t+1}^i-y_{t+1}^i)^2},
\end{equation}
where $\hat{y}_{t+1}^i$ and $y_{t+1}^i$ mean the prediction value and real value of region $i$ for time interval $t+1$, and where $\xi$ is total number of samples.
\subsection{Methods for Comparison}
We compared our model with the following methods, and tuned the parameters for all methods. We then reported the best performance.
\begin{itemize}
	\item \textbf{Historical average (HA)}: Historical average predicts the demand using average values of previous demands at the location given in the same relative time interval (i.e., the same time of the day).  
	\item \textbf{Autoregressive integrated moving average (ARIMA)}: ARIMA is a well-known model for forecasting time series which combines  moving average and autoregressive components for modeling time series.
	\item \textbf{Linear regression (LR)}: We compare our method with different versions of linear regression  methods: ordinary least squares regression (OLSR), Ridge Regression (i.e., with $\ell_2$-norm regularization), and Lasso (i.e., with $\ell_1$-norm regularization).
	\item \textbf{Multiple layer perceptron~(MLP)}: We compare our method with a neural network of four fully connected layers. The number of hidden units are $128$, $128$, $64$, and $64$ respectively.
	\item \textbf{XGBoost}~\cite{chen2016xgboost}: XGBoost is a powerful boosting tree based method and is widely used in data mining applications.
	\item \textbf{ST-ResNet}~\cite{zhang2016deep}: ST-ResNet is a deep learning based approach for traffic prediction. The method constructs a city's traffic density map at different times as images. CNN is used to extract features from historical images.
\end{itemize}
We used the same context features for all regression methods above. For fair comparisons, all methods (except ARIMA and HA) use the same loss function as our method defined in Eq.~\eqref{eq:loss}.

We also studied the effect of different view components proposed in our method.
\begin{itemize}
	\item \textbf{Temporal view}: For this variant, we used only LSTM with inputs as context features. Note that, if we do not use any context features but only use the demand value of last timestamp as input, LSTM does not perform well. It is necessary to use context features to enable LSTM to model the complex sequential interactions for these features.
	\item \textbf{Temporal view + Semantic view}: This method captures both temporal dependency and semantic information.
	\item \textbf{Temporal view + Spatial (Neighbors) view}: In this variant, we used the demand values of nearby regions at time interval $t$ as $\hat{\mathbf{s}}^i_t$ and combined them with context features as the input of LSTM. We wanted to demonstrate that simply using neighboring regions as features cannot model the complex spatial relations as our proposed local CNN method.
	\item \textbf{Temporal view + Spatial (LCNN) view}: This variant considers both temporal and local spatial views. The spatial view uses the proposed local CNN for considering neighboring relation. Note that when our local CNN uses a local window that is large enough to cover the whole city, it is the same as the global CNN method. We studied the performance of different parameters and show that if the size is too large, the performance is worse, which indicates the importance of locality. 
	\item \textbf{DMVST-Net}: Our proposed model, which combines spatial, temporal and semantic views.
\end{itemize}
\subsection{Preprocessing and Parameters}
We normalized the demand values for all locations to $[0,1]$ by using Max-Min normalization on the training set. We used one-hot encoding to transform discrete features (e.g., holidays and weather conditions) and used Max-Min normalization to scale the continuous features (e.g., the average of demand value in last four time intervals). As our method outputs a value in $[0,1]$, we applied the inverse of the Max-Min transformation obtained on training set to recover the demand value.

All these experiments were run on a cluster with four NVIDIA P100 GPUs. The size of each neighborhood considered was set as $9\times 9$ (i.e., $S$ = 9), which corresponds to $6km \times 6km$ rectangles. For spatial view, we set $K=3$ (number of layers), $\tau=3\times 3$ (size of filter), $\lambda=64$ (number of filters used), and $d=64$ (dimension of the output). For the temporal component, we set the sequence length $h=8$ (i.e., 4 hours) for LSTM. The output dimension of graph embedding is set as $32$. The output dimension for the semantic view is set to $6$. We used Sigmoid function as the activation function for the fully connected layer in the final prediction component. Activation functions in other fully connected layers are ReLU.  Batch normalization is used in the local CNN component. The batch size in our experiment was set to $64$. The first $90\%$ of the training samples were selected for training each model and the remaining $10\%$ were in the validation set for parameter tuning. We also used early-stop in all the experiments. The early-stop round and the max epoch were set to $10$ and $100$ in the experiment, respectively. 
\subsection{Performance Comparison}
\subsubsection{Comparison with state-of-the-art methods.} Table~\ref{tab:baseline} shows the performance of the proposed method as compared to all other competing methods. DMVST-Net achieves the lowest MAPE ($0.1616$) and the lowest RMSE ($9.642$) among all the methods, which is $12.17\%$ (MAPE) and $3.70\%$ (RMSE) relative improvement over the best performance among baseline methods. More specifically, we can see that HA and ARIMA perform poorly (i.e., have a MAPE of $0.2513$ and $0.2215$, respectively), as they rely purely on historical demand values for prediction. Regression methods (OLSR, LASSO, Ridge, MLP and XGBoost) further consider context features and therefore achieve better performance. Note that the regression methods use the same loss function as our method defined in Eq.~\eqref{eq:loss}. However, the regression methods do not model the temporal and spatial dependency. Consequently, our proposed method significantly outperforms those methods. 

Furthermore, our proposed method achieves $18.01\%$ (MAPE) and $6.37\%$ relative improvement over ST-ResNet. Compared with ST-ResNet, our proposed method further utilizes LSTM to model the temporal dependency, while at the same time considering context features. In addition, our use of local CNN and semantic view better captures the correlation among regions. 

\begin{table}[t]
	\begin{center}
		\caption{Comparison with Different Baselines}
		\begin{tabular}{l|c|c}
			\hline
			Method & MAPE  & RMSE\\\hline
			Historical average & 0.2513 & 12.167\\
			ARIMA & 0.2215 & 11.932\\
			Ordinary least square regression &  0.2063 &10.234\\
			Ridge regression & 0.2061 & 10.224\\
			Lasso & 0.2091& 10.327\\
			Multiple layer perceptron & 0.1840 & 10.609\\
			XGBoost& 0.1953 & 10.012\\
			ST-ResNet & 0.1971 & 10.298\\\hline
			DMVST-Net & \textbf{0.1616} & \textbf{9.642}\\\hline
		\end{tabular}
		\label{tab:baseline}
	\end{center}
\end{table}
\subsubsection{Comparison with variants of our proposed method.} Table~\ref{tab:variants} shows the performance of DMVST-Net and its variants. First, we can see that both Temporal view + Spatial (Neighbor) view and Temporal view + Spatial (LCNN) view achieve a lower MAPE (a reduction of $0.63\%$ and $6.10\%$, respectively). The result demonstrates the effectiveness of considering neighboring spatial dependency. Furthermore, Temporal view + Spatial (LCNN) view outperforms Temporal view + Spatial (Neighbor) view significantly, as the local CNN can better capture the nonlinear relations. On the other hand, Temporal view + Semantic view has a lower MAPE of $0.1708$ and an RMSE of $9.789$ compared to Temporal view only, demonstrating the effectiveness of our semantic view. Lastly, the performance is best when all views are combined. 
\begin{table}[t]
	\begin{center}
		\caption{Comparison with Variants of DMVST-Net}
		\begin{tabular}{l|c|c}
			\hline
			Method & MAPE  & RMSE\\\hline
			Temporal view & 0.1721 & 9.812\\
			Temporal + Semantic view & 0.1708 & 9.789\\
			Temporal + Spatial (Neighbor) view & 0.1710 & 9.796\\
			Temporal + Spatial (LCNN) view & 0.1640 & 9.695\\
			DMVST-Net & \textbf{0.1616} & \textbf{9.642}\\\hline
		\end{tabular}
		\label{tab:variants}
	\end{center}
\end{table}
\subsection{Performance on Different Days}
\begin{figure}[t]
	\centering
	\includegraphics[width=2.8in]{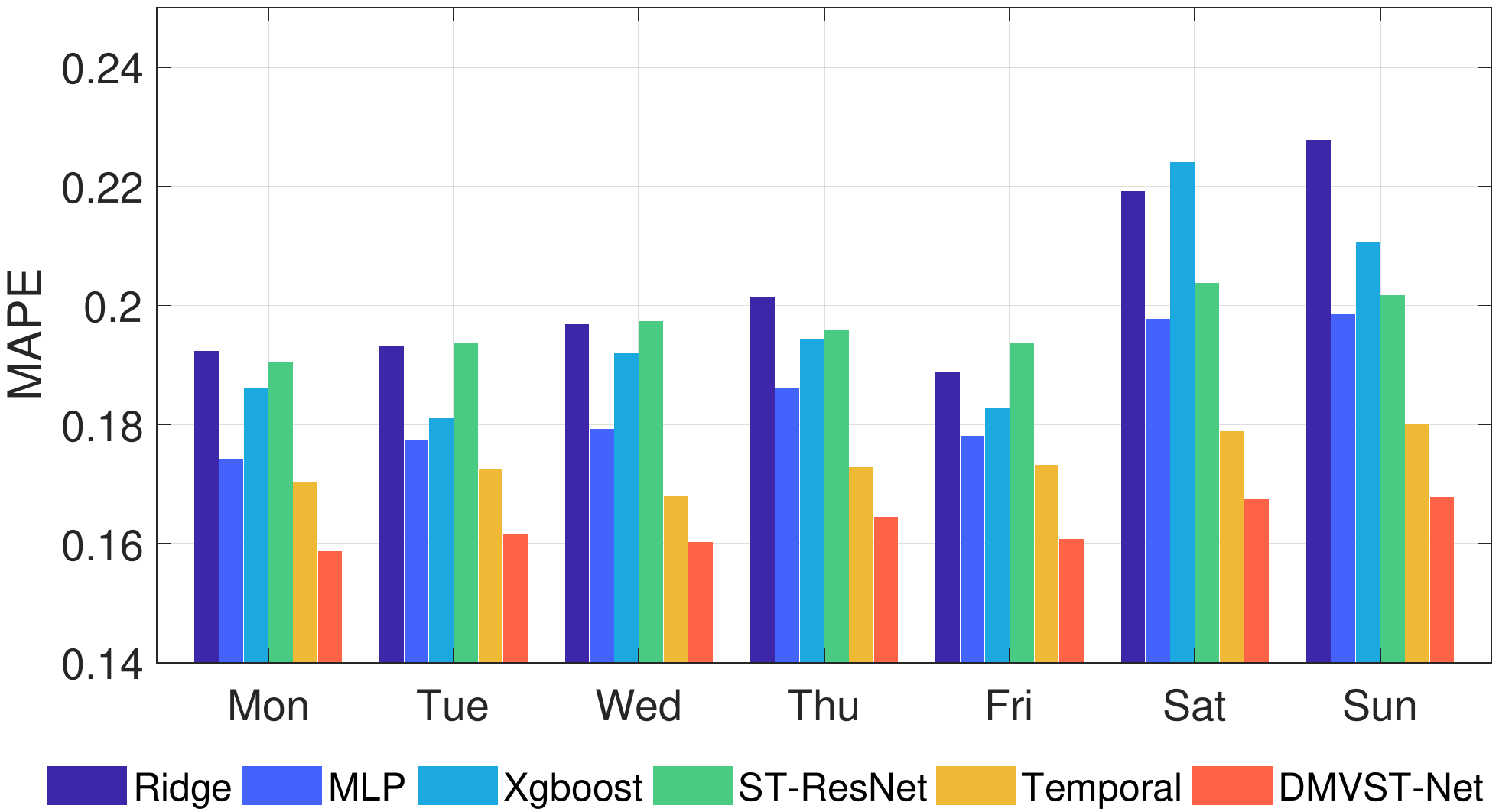}
	\caption{The Results of Different Days.}
	\label{fig:day}
\end{figure}

Figure~\ref{fig:day} shows the performance of different methods on different days of the week. Due to the space limitation, We only show MAPE here. We get the same conclusions of RMSE. We exclude the results of HA and ARIMA, as they perform poorly. We show Ridge regression results as they perform best among linear regression models. In the figure, it shows that our proposed method DMVST-Net outperforms other methods consistently in all seven days. The result demonstrates that our method is robust. 

Moreover, we can see that predictions on weekends are generally worse than on weekdays. Since the average number of demand requests is similar ($45.42$ and $43.76$ for weekdays and weekends, respectively), we believe the prediction task is harder for weekends as demand patterns are less regular. For example, we can expect that residential areas may have high demands in the morning hours on weekdays, as people need to transit to work. Such regular patterns are less likely to happen on weekends. To evaluate the robustness of our method, we look at the relative increase in prediction error on weekends as compared to weekdays, i.e., defined as $|\bar{wk}-\bar{wd}|/\bar{wd}$, where $\bar{wd}$ and $\bar{wk}$ are the average prediction error of weekdays and weekends, respectively. The results are shown in Table~\ref{tab:day}. For our proposed method, the relative increase in error is the smallest, at $4.04\%$.

At the same time, considering temporal view, only (LSTM) has a relative increase in error of $4.77\%$, while the increase is more than $10\%$ for Ridge regression, MLP, and XGBoost. The more stable performance of LSTM can be attributed to its modeling of the temporal dependency. We see that ST-ResNet has a more consistent performance (relative increase in error of $4.41\%$), as the method further models the spatial dependency. Finally, our proposed method is more robust than ST-ResNet.

\begin{table}[t!]
	\begin{center}
		\caption{Relative Increase in Error (RIE) on Weekends to Weekdays}
		\scriptsize
		\begin{tabular}{@{\hskip 0.04in}c@{\hskip 0.04in}|@{\hskip 0.04in}c@{\hskip 0.04in}|@{\hskip 0.04in}c@{\hskip 0.04in}|@{\hskip 0.04in}c@{\hskip 0.04in}|@{\hskip 0.04in}c@{\hskip 0.04in}|@{\hskip 0.04in}c@{\hskip 0.04in}|@{\hskip 0.04in}c@{\hskip 0.04in}}
			\hline
			Method                                                       & RIDGE   & MLP     & XGBoost & ST-ResNet & Temporal & DMVST-Net \\ \hline
			\begin{tabular}[c]{@{}l@{}}RIE\end{tabular} & 14.91\% & 10.71\% & 16.08\% & 4.41\%    & 4.77\%   & \textbf{4.04\%}    \\ \hline
		\end{tabular}
		\label{tab:day}
	\end{center}
\end{table}

\subsection{Influence of Sequence Length for LSTM}
In this section, we study how the sequence length for LSTM affects the performance. Figure~\ref{fig:lstmlen} shows the prediction error of MAPE with respect to the length. We can see that when the length is $4$ hours, our method achieves the best performance. The decreasing trend in MAPE as the length increases shows the importance of considering the temporal dependency. Furthermore, as the length increases to more than $4$ hours, the performance slightly degrades but mainly remains stable. One potential reason is that when considering longer temporal dependency, more parameters need to be learned. As a result, the training becomes harder.

\subsection{Influence of Input Size for Local CNN}
\begin{figure}[t!]
	\centering
	\begin{subfigure}[b]{0.22\textwidth}
		\centering
		\includegraphics[height=0.8\textwidth]{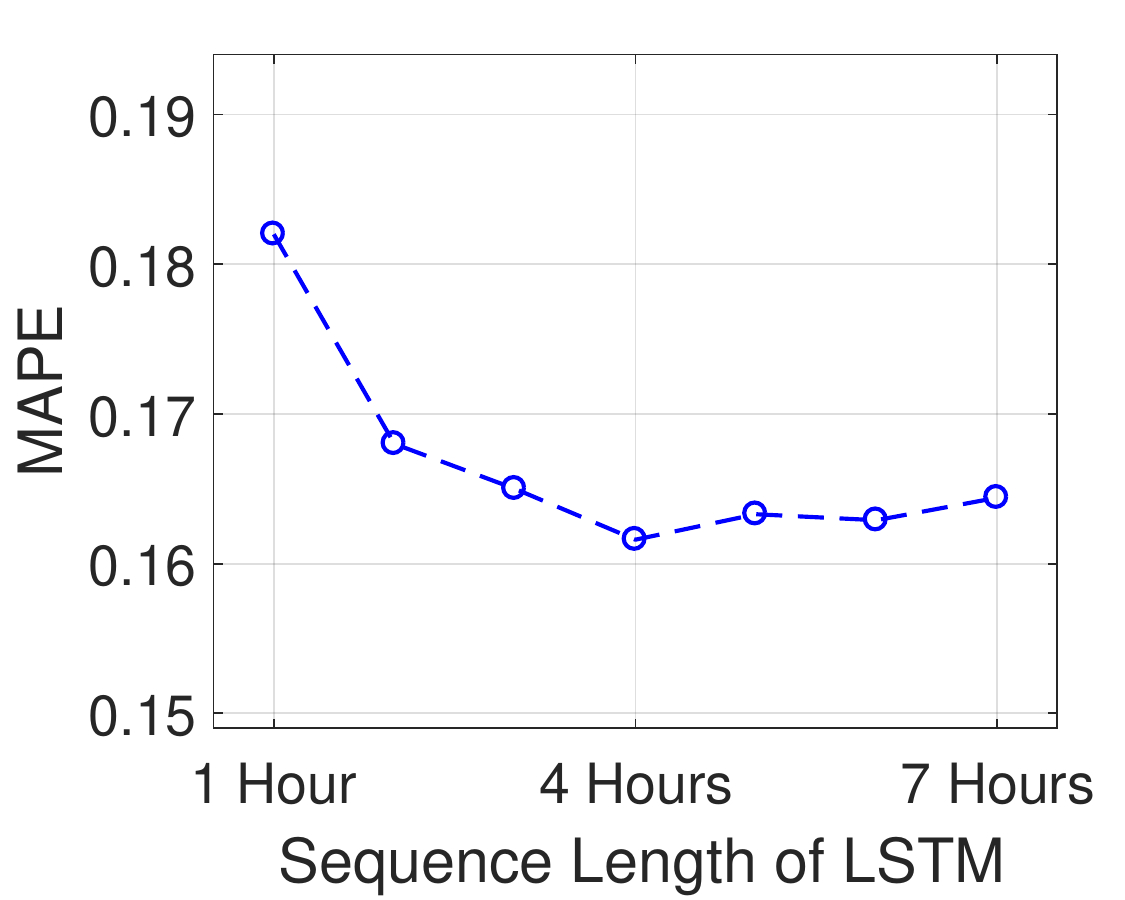}
		\caption{\label{fig:lstmlen}}
	\end{subfigure}
	\begin{subfigure}[b]{0.22\textwidth}
		\centering
		\includegraphics[height=0.8\textwidth]{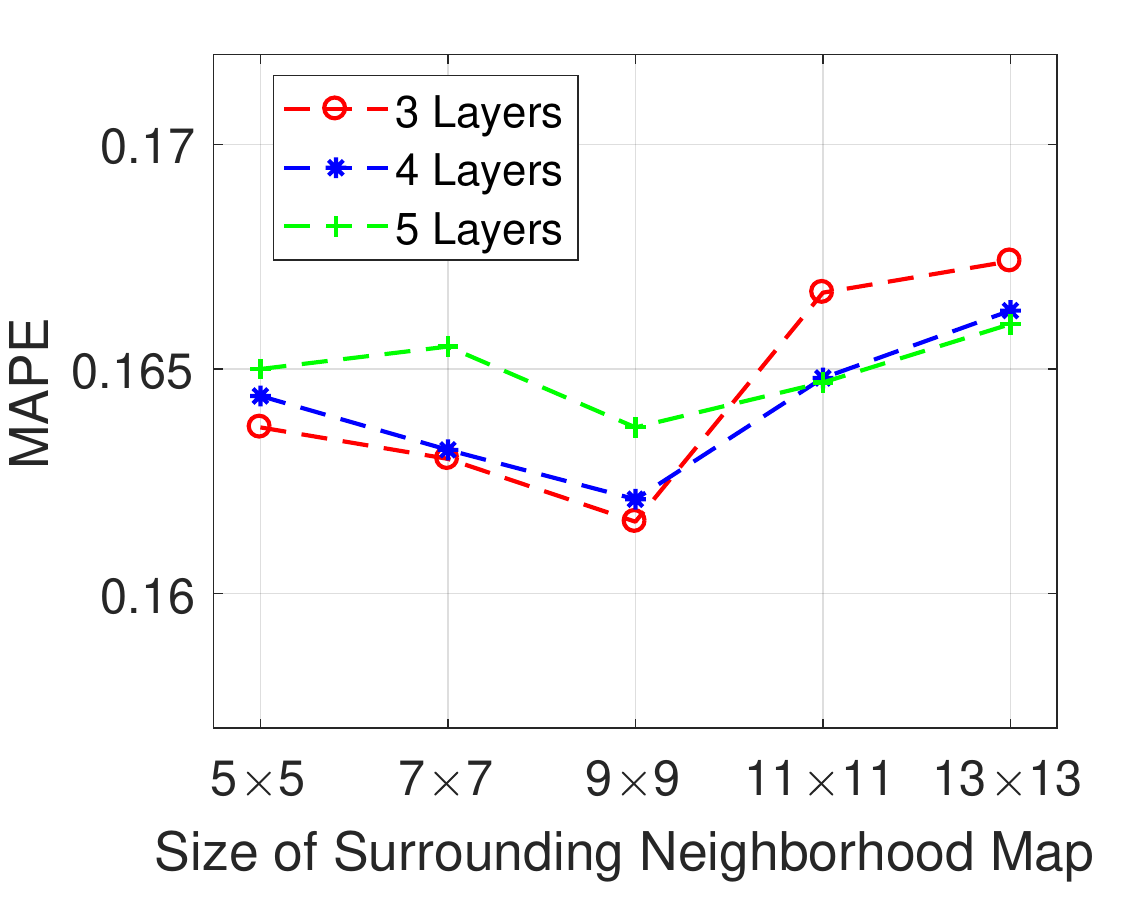}
		\caption{\label{fig:depthsize}}
	\end{subfigure}
	\caption{(\subref{fig:lstmlen}) MAPE with respect to sequence length for LSTM. (\subref{fig:depthsize}) MAPE with respect to the input size for local CNN.}
\end{figure}

Our intuition was that applying CNN locally avoids learning relation among weakly related locations. We verified that intuition by varying the input size $S$ for local CNN. As the input size $S$ becomes larger, the model may fit for relations in a larger area. In Figure~\ref{fig:depthsize},  we show the performance of our method with respect to the size of the surrounding neighborhood map. We can see that when there are three convolutional layers and the size of map is $9\times 9$, the method achieves the best performance. The prediction error increases as the size decreases to $5 \times 5$. This may be due to the fact that locally correlated neighboring locations are not fully covered. Furthermore, the prediction error increases significantly (more than $3.46\%$), as the size increases to $13 \times 13$ (where each area approximately covers more than $40\%$ of the space in GuangZhou). The result suggests that locally significant correlations may be averaged as the size increases. We also increased the number of convolution layers to four and five layers, as the CNN needed to cover larger area. However, we observed similar trends of prediction error, as shown in Figure~\ref{fig:depthsize}. We can now see that the input size for local CNN when the method performs best remains consistent (i.e., the size of map is $9\times 9$).

\section{Conclusion and Discussion}
The purpose of this paper is to inform of our proposal of a novel Deep Multi-View Spatial-Temporal Network (DMVST-Net) for predicting taxi demand. Our approach integrates the spatial, temporal, and semantic views, which are modeled by local CNN, LSTM and semantic graph embedding, respectively. We evaluated our model on a large-scale taxi demand dataset. The experiment results show that our proposed method significantly outperforms several competing methods.  

As deep learning methods are often difficult to interpret, it is important to understand what contributes to the improvement. This is particularly important for policy makers. For future work, we plan to further investigate the performance improvement of our approach for better interpretability. In addition, seeing as the semantic information is implicitly modeled in this paper, we plan to incorporate more explicit information (e.g., POI information) in our future work.

\section{Acknowledgments}
The work was supported in part by NSF awards \#1544455, \#1652525,
\#1618448, and \#1639150. The views and conclusions contained in
this paper are those of the authors and should not be interpreted
as representing any funding agencies.
\bibliographystyle{aaai}
\fontsize{9.0pt}{10.0pt} \selectfont
\bibliography{ref}
\end{document}